Preprint of a paper published in Synthese 206: 115, 2025 (creative commons licence)

# Can AI systems have free will?

Christian List*

This version: August 2025, to appear in *Synthese*

**Abstract:** While there has been much discussion of whether AI systems could function as moral agents or acquire sentience, there has been very little discussion of whether AI systems could have free will. I sketch a framework for thinking about this question, inspired by Daniel Dennett's work. I argue that, to determine whether an AI system has free will, we should not look for some mysterious property, expect its underlying algorithms to be indeterministic, or ask whether the system is unpredictable. Rather, we should simply ask whether we have good explanatory reasons to view the system as an intentional agent, with the capacity for choice between alternative possibilities and control over the resulting actions. If the answer is "yes", then the system counts as having free will in a pragmatic and diagnostically useful sense.

## 1. Introduction

The increasing use of AI systems in society raises many questions: Are such systems safe? Do they behave ethically? Should they be given a legal status of their own, and who is responsible when they cause harms? Could AI systems even become conscious?[1] But despite the explosion of work on these and related questions, one question has received surprisingly little attention: can AI systems have free will? Free will is associated with autonomous agency and is often considered a precondition for moral responsibility. The question of free will thus matters for debates about whether AI systems could count as responsible agents in their own right. Perhaps the thought that free will is distinctively human has led many commentators to ignore the possibility of artificial free will.

In this paper, I will discuss what it would take for an AI system to have free will. To determine whether a system has free will, I will argue, we should assess the system by reference to a checklist of three conditions: "intentional agency", "alternative possibilities", and "causal control" (List 2019). If a system meets all three conditions, then it qualifies as having free will in a practically useful sense. On this basis, I will argue that free will in AI is much less far-fetched than perhaps expected. That said, my aim is not so much to settle the question of whether any current AI system has free will but rather to set out the criteria for free will in AI. My analysis is inspired by Daniel Dennett's work (especially his work on the "intentional stance", but also his work on free will; e.g., Dennett 1984, 1987, 2003). However, my analysis goes beyond Dennett's, among other things by putting a stronger emphasis on the idea that free will requires the possibility of doing otherwise (following List 2019).

---

* Munich Center for Mathematical Philosophy, LMU Munich, Munich, Germany, c.list@lmu.de

[1] For discussion, see, among many others, Chalmers (2023), Delcker (2018), Dubber, Pasquale, and Das (2020), Floridi and Sanders (2004), Floridi (2023), Fossa (2018), Matthias (2004), Nyholm (2020), and Solum (1992).

Among the few prior works discussing free will in AI, most reach more negative conclusions, in part because they employ conditions for free will that are more demanding than the ones I will defend here (see, e.g., Floridi and Sanders 2004, Krausová and Hazan 2013, Farnsworth 2017, Sanchis 2018, and Blum and Blum 2022).[2] Other works focus on the public's *beliefs* about whether AI systems have free will (Astobiza 2024), and these beliefs also suggest a broadly negative answer. The closest precursor to the present analysis can be found in Maier (2023), who, like me, argues for the possibility of free will in AI by pointing out that AI systems can be viewed as choice-making agents to which decision-theoretic models can be applied. I will develop this idea in terms of the above-mentioned checklist of three conditions, whose application to AI was already briefly anticipated in List (2019, pp. 156–157) and further discussed in a recent paper by Martela (2025).[3] The proposed checklist is meant to capture a commonsense understanding of free will that is neither too weak nor unrealistically demanding. It is consistent with Dennett's idea that we should think of free will, not as requiring any mysterious metaphysical powers, but as a capacity that is part and parcel of the kind of agency we find in humans and other complex animals – a capacity that enables them to navigate their environments in flexible and adaptive ways.

## 2. What is AI?

Artificial intelligence can be defined as the capacity of an artificial system, such as a computational or robotic one, to perform cognitive tasks and/or to interact with the environment in ways traditionally associated with human or animal intelligence. In line with this definition, the academic field of AI has been characterized as "the study of agents that receive percepts from the environment and perform actions" (Russell and Norvig 2021, p. 7).

One common distinction is that between "weak" and "strong" AI. "Weak AI" refers to artificial intelligence narrower than human intelligence, for instance due to being restricted to fixed computational tasks, while “strong AI” refers to artificial intelligence more similar to human intelligence in its flexibility or generality. “Artificial general intelligence”, moreover, refers to a form of artificial intelligence on a par with or stronger than human intelligence across many tasks. Examples of weak AI systems include chess-playing computers and smart route planners. Strong AI is increasingly becoming a reality, as exemplified by generative AI chatbots that can conduct complex conversations or compose texts on many subjects. The quest for artificial general intelligence is the industry’s next frontier, although there is no consensus on how close we are to achieving it and how desirable it would be. While some commentators, such as Luciano Floridi (2023), have suggested that current AI systems are best viewed as displaying a new form of agency without any human-like intelligence, others, such as Blaise Agüera y Arcas and Peter Norvig (2024), think that “decades from now, [today’s most advanced AI large language models] will be recognized as the first true examples of artificial general intelligence”. Those systems, they note, flexibly cover many topics, perform many tasks, across different languages, can process not just images and text, but also audio and video, are connectable to robotic devices, and have advanced learning capacities.[4] Indeed, current AI systems already make or participate in decisions that used to be the exclusive domain of humans, whether it is driving decisions in transportation, diagnostic decisions in medicine, financial decisions in banking and investment, juridical decisions in legal contexts, or targeting decisions in the military.

[2] Floridi and Sanders (2004) require human-like internal states for free will and argue that "there is substantial and important scope for the concept of moral agent not necessarily exhibiting free will or mental states" (p. 351). Farnsworth (2017) suggests that "[t]he main impediment to free-will in present-day artificial robots, is their lack of being a Kantian whole" (p. 1). More congenially from my perspective, he emphasizes the importance of choice-making for free will. Both Krausová and Hazan (2013) and Sanchis (2018) relate free will to unpredictability. As explained in section 5.3 below, I do not think that unpredictability is required for free will. Blum and Blum (2022) define free will as "the ability to violate physics" (p. 2) and suggest that "[a]ll theory is against the freedom of the will" (p. 3). However, by appealing to the concept of a "Conscious Turing Machine (CTM)", they argue that "[a]lthough a deterministic entity living in a deterministic world cannot have free will as it is typically understood, it is entirely possible for such an entity to rightly and firmly *believe* in its own free will" (p. 1, emphasis added). More congenially, they define "the exercise of free will" as "the act of choosing between two or more options" (p. 2) and suggest that "[t]his is something that any animal or machine with a CTM brain can do" (ibid.).

[3] Martela's paper (2025), published after I submitted the present paper, also draws on List (2019) to defend the possibility of free will in AI, arguing that "generative agents utilizing large language models have functional free will". While Martela's approach is broadly consistent with mine, my focus, as noted, is not on the claim that current AI systems already have free will but rather on developing a framework for thinking about artificial free will. One respect in which I differ from Martela is that I wouldn't draw a distinction between what he calls "physical free will" and "functional free will", since it may inadvertently create the impression that "functional free will" is a lesser form of free will than "physical free will", when in fact free will, as I understand it, is exclusively an agential-level notion rather than a physical one. I would just speak of free will simpliciter.

It is tempting to characterize AI by reference to the underlying technology. “Symbolic AI”, the dominant approach in the second half of the twentieth century, refers to the implementation of AI through the explicit processing of symbolic representations, using tools from logic. “Subsymbolic AI”, which is now dominant, refers to the implementation of AI through some lower-level architecture, such as a neural network, which does not by itself come with a symbolic interpretation. The leading version of this approach, “generative AI”, implements AI by means of machine-learning algorithms that can generate new content, prompted by certain inputs, on the basis of statistical patterns picked up from training data. However, while the technology is important, I suggest that the *definition* of AI should focus on the cognitive and agentive capacities achieved rather than on the precise technology used to achieve it.

## 3. What is free will?

Free will can be defined as “a particular sort of capacity of rational agents to choose a course of action from among various alternatives” (O’Connor 2010; for an overview, see Kane 2002). According to our commonsense understanding, we human beings have this capacity. It is your own free choice, for instance, to read this paper or to refrain fom reading it. If you have started reading it, it is up to you whether to continue or stop.

What does free will require? Free will is sometimes characterized, especially by sceptics, in ways that make it seem mysterious (see, e.g., Harris 2012, Sapolsky 2023). For instance, if free will is taken to require the ability to make “contra-causal” choices – choices that are unconstrained by the laws of nature or that even overrule them – then free will seems in conflict with a scientific worldview. Similarly, if free will is taken to require control, not just over one's actions but also over their entire causal pre-history, including everything that has shaped one's personality, preferences, and beliefs, then free will is also impossible.

---

[4] For discussion, see also Ananthaswamy (2024).

In real life, however, we are not normally interested in any such unrealistically strong capacities. The "varieties of free will worth wanting", as Dennett (1984) calls them, don't involve such capacities, and commonsense reasoning doesn't postulate them either. When we ask in everyday contexts whether people choose and control their actions or when we assess people's responsibility for their actions, we simply refer to the kinds of ordinary agential competences that humans normally acquire in their pathway towards becoming competent adults. When a judge takes someone's free will to be a precondition for legal responsibility – for anything ranging from a breach of contract to criminal conduct – the judge does not require that the person can break the laws of nature or had control over their entire character-forming process (see Moore 2020). That would rule out legal responsibility from the outset and rob the idea of free will of its practical usefulness. Rather, the judge is concerned with the ordinary exercise of choice and control that we attribute to each other in commonsense psychology.

We distinguish between a premeditated crime, committed by someone in full possession of their cognitive and agentive capacities, and an accidental harm caused by a sleep-walker. In the premeditated case, judges attribute the action to the person's free will; in the sleep-walking case, they don't. Similarly, we distinguish between a competent adult whose ability to act is not physiologically or psychologically compromised and someone who is intoxicated or acts out of compulsion. The former is held responsible for their action, the latter not. A useful understanding of free will – the kind of "free will worth wanting", as Dennett calls it – should support those distinctions and not rule out free will from the outset. Free will, Dennett (2003) suggests, is a biologically evolved capacity that is part and parcel of human agency. It enables humans to navigate complex environments in a flexible and adaptive way.

Consistently with these ideas, I find it helpful to define free will in terms of three conditions (following List 2019):

**Intentional agency:** Any bearer of free will is an intentional agent, i.e., an entity capable of acting in a goal-directed manner, based on intentional states such as beliefs and desires.

**Alternative possibilities:** Any bearer of free will has alternative possibilities to choose from, i.e., different courses of action this entity could take.

**Causal control:** Any bearer of free will has relevant control over the actions taken, in the sense that the entity's intentional states are the difference-making causes of those actions.

I will assume that these three conditions – perhaps with some fine-tuning – are jointly necessary and sufficient for free will. That is: whenever someone or something qualifies as an intentional agent, has the ability to choose between alternative possibilities, and has sufficient control over the resulting actions, he, she, or it counts as having free will. If one or more of these conditions are violated, then not. The distinctions drawn by judges illustrate the present conditions.

Someone in full possession of their cognitive and agentive capacities normally meets them; someone sleep-walking does not. A competent adult signing a contract in normal circumstances presumably meets all three conditions; someone who acts out of compulsion or while intoxicated does not.

Some philosophers, including Dennett in some of his writings (e.g., 1984, ch. 6), suggest that alternative possibilities are not needed for free will and that less demanding conditions suffice (see also Frankfurt 1969 and the overview in Kane 2002). The idea is that free will is compatible with the lack of alternative possibilities as long as the agent genuinely intends their action and appropriately qualifies as its "author". Dennett (1984, ch. 6) gives the example of Martin Luther, the church reformer in the 16th century, who, when asked to either reaffirm or renounce his criticism of the Roman Catholic Church, reaffirmed that criticism and allegedly said "Here I stand; I can do no other". Dennett notes that we regard this as a free choice on Luther's part, despite his apparent inability to act otherwise. After all, Luther "authored" his action, even if he could not have acted otherwise. However, contrary to this interpretation, I think that Luther should not be regarded as having been literally incapable of acting otherwise. Rather, he could not have acted otherwise without abandoning his values, and he was not prepared to do that. He had a genuine choice, albeit one in which he strongly endorsed one of the options and rejected the other (for discussions of this example on which I draw, see also Kane 2002, ch. 1, List 2014, and List and Rabinowicz 2014).

For this reason, I side with those who think that free will requires alternative possibilities. This position is held by so-called "free-will libertarians", who think that there could not be any free will without a form of indeterminism or openness of the future, but it is also held by those "compatibilists" who define the notion of "having alternative courses of action" in a way that is compatible with determinism.[5] Indeed, Dennett, in other writings, suggests that there is a sense in which determinism does not imply "inevitability" (Dennett 2003), and that there is a "broad" interpretation of "possibility" relative to which an agent could be said to have alternative possibilities even in a deterministic world (Taylor and Dennett 2002). (More on this in section 5.2 below.) And the theory developed in List (2014, 2019) seeks to back this up further by introducing a notion of "agential possibility" relative to which choice-making agents can be said to have alternative possibilities (see also Maier 2015, 2022).

The insistence on alternative possibilities as a requirement for free will is arguably consistent with common sense, which tends to represent free choices as requiring a "fork in the road", where the agent could do one thing or another (e.g., Carstensen et al. 2025). By taking free will to require alternative possibilities, then, the present understanding of free will can defend itself against the charge that it is too "watered down" – a charge that is sometimes levelled against those understandings of free will that give up alternative possibilities (see, e.g., Kane 2007, p. 180). This also sets my account of free will apart from the version of Dennett's account that drops that requirement (e.g., in 1984, ch. 6).

[5] For a discussion of the available strategies, see List (2014).

Free-will sceptics claim that people never have free will in the sense required here. What we conventionally consider a free choice, the sceptics say, is no more under our control than bodily reflexes or compulsions. For the sceptics, free will is an illusion: everything is the consequence of a physical system inexorably evolving under the laws of nature (Pereboom 2001, Sapolsky 2023). This paper is not the place to respond to such free-will scepticism in general. As Dennett (2003) has pointed out, however, it is plausible to expect free will to have evolved under the pressure to survive in complex environments, insofar as the flexible decision-making capacities that come with free will are clearly evolutionarily advantageous. Furthermore, society, including our legal system, tends to assume that human beings normally have free will, in the sense of possessing its three constituent capacities: agency, choice between alternative possibilities, and control over the resulting actions. The sciences of human behaviour operate on this assumption, too (List 2019, 2023). Disciplines ranging from economics and political science to anthropology and history all take what Dennett (1987) calls "an intentional stance" towards the people they study, assuming that *they are intentional agents who make choices and have a relevant form of causal control over those choices*. Those sciences thereby presuppose that people have free will under the present definition. Call this the "free-will presupposition".

If we wanted to eliminate this presupposition from our explanations of human behaviour, we would need to change our entire explanatory paradigm: all references to agency and choice would have to be replaced by references to external factors and impersonal causal processes. We would have to abandon the "intentional stance" in favour of a "physical stance". We would no longer be talking about agents making choices. Rather, we would have to reconceptualize people as mere physical systems or passive spectators: organisms that are moved by factors beyond their control, just as air molecules float around in a thermodynamic system. *Intentional explanations*, which depict people as goal-directed agents who make intelligible choices between different possible actions, would have to be replaced by *dynamic or stochastic explanations*, along the lines of how we explain the motion of the planets, heat diffusion, or fluid dynamics. The fact that intentional explanations are so central to so many explanatory practices – from commonsense psychology and our legal system to the human and social sciences – lends further support to free will as a working hypothesis (List 2019, 2023).

## 4. Free will in AI

It may seem far-fetched to look for free will in a system that is, at bottom, nothing more than a digital computer interacting with its environment. Indeed, if free will required some mysterious property such as an inner "homunculus" or the ability to transcend the laws of nature, then AI systems would be unlikely candidates for having free will. However, so would humans. As Dennett (2003, pp. 2-3) notes,

> "What you are is an assemblage of roughly a hundred trillion cells, of thousands of different sorts. … Each of your host cells is a mindless mechanism, a largely autonomous micro-robot. … The more we learn about how we have evolved, and how our brains work, the more certain we are becoming that there is no … extra ingredient. We are each made of mindless robots and nothing else, no non-physical, non-robotic ingredients at all."

Thus, if we are prepared to say that human beings – as composite systems consisting of seemingly mindless building blocks – have free will, then we must also accept that the underlying mechanical make-up of an AI system does not by itself rule out free will in such a system. On the analysis proposed here, to determine whether an AI system has free will, we must use the checklist of the three above-stated conditions to assess the system: an AI system has free will *if and only if* it has intentional agency, alternative possibilities to choose from, and causal control over its actions (List 2019). So, I propose to use the same criteria for identifying free will in AI that we also use in the case of humans or biological animals.

How do we establish whether a given system meets those conditions? Here, Dennett's methodology from his work on the "intentional stance" is helpful, though I will slightly reinterpret Dennett's approach. To determine whether a given system is an intentional agent, Dennett suggests, we should not get too carried away with deep metaphysical questions but ask whether the system can be *well-explained by viewing it as an intentional agent*. For Dennett (2009, p. 339), "[a]nything that is usefully and voluminously predictable from the intentional stance is, by definition, an *intentional system*". On this account, to be an intentional agent is simply to be a system that can be well-explained by taking an intentional stance towards it. In a similar vein, one might propose, anything that is well-explained by viewing it as a system with intentional agency, alternative possibilities, and causal control over the resulting actions should count as having free will.

Dennett's account, in its core form, suggests that whether a system *is* of a certain kind boils down to whether the system is *interpretable* (or *well interpretable*) as being of that kind. This makes it seem as if a capacity such as agency is largely in the eye of the beholder: anything an interpreter can "usefully and voluminously" predict by viewing it as an agent is "by definition" an agent. Many people will find this too interpretivist. I will therefore re-interpret Dennett's intentional-stance criterion in a more realist manner. According to the desired realist account, a system's capacities are real and not just in the eye of the beholder. We can then treat the availability of a certain kind of explanation of the system, such as an intentional-stance explanation, as an *indicator* of the system's capacities, not as the *defining criterion* for those capacities.[6] If a system can be well-explained from the intentional stance, and especially if taking the intentional stance is explanatorily necessary, this is *evidence* that the system is really an agent. Compare the following: the fact that physical systems can be well-explained (or best-explained) by assuming that there is gravity and electromagnetism is *evidence* for the hypothesis that there really is gravity and electromagnetism.

In line with this, I propose that, to determine whether a given system has free will, we should ask whether we have *good explanatory reasons* to view the system as an entity that meets the three above-mentioned conditions, i.e., as (i) an intentional agent with (ii) alternative possibilities to choose from and (iii) control over its actions. I will now run through these conditions and explain what would be required for a positive answer.

---

[6] Note that an *indicator* for something need not be the *defining criterion* for it. For instance, my possession of a driving licence is an indicator of my ability to drive, but it is not the defining criterion of that ability; it is not constitutive of that ability. For me, the explanatory indispensability of the intentional stance in relation to a given system is *indicative* of the system's intentional agency, not *constitutive* of it. Dennett, by contrast, sometimes uses the availability or indispensability of the intentional stance as the defining criterion of agency.

*4.1. Intentional agency*

An *intentional agent* is an entity capable of acting in a goal-directed manner, based on intentional states such as beliefs and desires. Following a long-standing tradition in philosophy and computer science, we can define *beliefs* as representations of what things are like and *desires* as representations of a target state of things to be achieved or as rankings of different such states (or assignments of utility to them).[7] Given these definitions, we have good reasons to view many AI systems as agents with beliefs and desires. If we didn't take an intentional stance towards those systems and focused solely on their low-level algorithms, we wouldn't adequately capture their cognitive and agential capacities. Recall that Russell and Norvig characterize AI in terms of agency. AI systems, they remark, "operate autonomously, perceive their environment, persist over a prolonged time period, adapt to change, and create and pursue goals" (2001, p. 4).[8] Indeed, the agentive and cognitive capacities of AI systems have advanced dramatically in recent years, and more and more such systems seem to warrant an agential description.

That said, there is a debate on whether attributions of beliefs and other intentional states to AI systems are genuinely justified. It may be objected that the surface-level outputs of AI systems cannot generally be interpreted as accurate reflections of stable internal states that play a belief role. For example, a large language model may give inconsistent responses to different users or in response to different prompts, and it may merely function as a "stochastic parrot" (Bender et al. 2021): its outputs statistically fit its inputs, relative to some loss function. This would speak against the view that the system genuinely has the kinds of belief-and-desire states that are required for intentional agency.

However, as Levinstein and Herrmann (2025) have argued, "our best theories of belief and decision making make it a very live possibility that LLMs *do* have beliefs, since beliefs might very well be helpful for making good predictions about tokens" (p. 1543). Even if the underlying algorithm simply leads the system to produce outputs that minimize some loss function, without reference to anything like belief, truth, consistency, or representation, some internal states of the system could still play the role of representations or beliefs, as a byproduct of this minimization exercise. Levinstein and Herrmann note: "[i]t is easy to generate decision contexts (such as strategic board games, investing, figuring out how to get to Toronto from Prague, etc.) that do seem to push us [humans] to form accurate beliefs about the world" (p. 1545). Similarly, for many problem-solving tasks, AI systems would benefit from having internal states that play a belief role. As in the human case, "it is very useful to have an accurate map of the world, in order to guide action" (ibid., p. 1546). This suggests that it is "largely an empirical matter" whether AI systems have beliefs and other stable intentional states (ibid.).

[7] On belief-desire agency, see, e.g., Bratman (1987).

[8] The intentional stance is also evident in Russell and Norvig's more detailed taxonomy of the capacities of AI systems, where they distinguish between *thinking* humanly and/or rationally and *acting* humanly and/or rationally.

Some AI researchers have further suggested that AI systems develop "world models", i.e., the sorts of representations of the environment often associated with intentional agency (cf. Ananthaswamy 2024). Gurnee and Tegmark (2023), for instance, studied LLM systems trained on spatial and temporal datasets, containing data about the world, the United States, New York City, historical figures, news headlines, and so on. They concluded that the systems learned representations of space and time that were "robust to prompting variations and unified across different entity types (e.g. cities and landmarks)" (p. 1). They further found "individual 'space neurons' and 'time neurons' that reliably encode spatial and temporal coordinates", and cautiously suggested that "modern LLMs learn rich spatiotemporal representations of the real world and possess basic ingredients of a world model" (ibid.).

Now a critic might still say: perhaps it is a *useful heuristic* to talk *as if* AI systems were intentional agents, but in reality they are just mechanistic devices. Their algorithms leave no room for real intentional agency. However, if we take Dennett's intentional-stance criterion seriously, then the usefulness (perhaps even the explanatory indispensability) of ascribing agency to those systems should be at least good evidence that they really are intentional agents.

Furthermore, the present criticism fails to acknowledge that there are different levels at which we may describe an AI system. The availability of a low-level algorithmic description does not make a high-level description in terms of agency incorrect. To see this, first consider the case of a human being. At some level, the human organism is a biophysical system, where physical and chemical processes take place, neurons get activated, and electrochemical signals get transmitted. This may be the right level of description for some medical interventions and the neuroscientific study of brain processes. But for other purposes, it is essential to describe humans as intentional agents. The ascription of agency to people allows us to explain why they vote the way they do, why they show up for work, why they sometimes keep their promises, and so on. Similarly, different levels of description are available for computational systems. At the hardware level, we may view a computer as a physical system in which electricity flows through microchips, which in turn can be interpreted as the execution a large number of binary logical operations. At a higher level, however, we may view the computer as running intelligible software applications. No software engineer could dispense with that level of description. In the case of AI, in particular, there is not just a low level at which we focus on mechanisms or algorithms, but, to capture a system's cognitive or agentive performance, we may need to view the system as a goal-directed agent that responds intelligibly to its environment (see also Floridi and Sanders 2004). We may understand the system's functioning better by recognizing its high-level representations – its "beliefs" and "goals" – than by merely unpacking the workings of its algorithms. Taking an "intentional stance" may be explanatorily necessary.

In fact, the desideratum of explainability might directly favour the design of AI systems that qualify as intentional agents. A common criticism of AI is that the underlying mechanisms are opaque and hard to understand and predict. The quest for "explainable" AI is precisely the quest for intelligible explanations of why an AI system behaves the way it does. Explainability has been defined as "an approach focusing on providing understandable justifications for the decisions made by an AI model, aiming to answer questions like '*Why did the model make this decision?*'", where the explanation should be intelligible "from a user's perspective" (Retzlaff et al. 2024, p. 2; see also Miller 2019).[9] Explainability, so defined, may be hard to achieve at a low, algorithmic level of description, just as it is hard to explain human behaviour by looking exclusively at neurons firing while ignoring high-level cognitive and agentive functions.[10] The quest for explainability may give us reasons to design AI systems whose functioning is comprehensible in agential terms. It may prompt us to look for system architectures that support stable belief-like and goal-like states that render the system's behaviour intelligible. Questions about why a system performed certain actions are often best answered by giving high-level explanations, such as intentional ones (see also Miller 2019).[11]

*4.2. Alternative possibilities*

If, as I have argued, we have good reasons to view some AI systems as intentional agents, we must then ask whether we have good reasons to think that those systems have alternative possibilities to choose from. In fact, it can also be argued that once we explain an entity's behaviour by viewing it as an intentional agent, we must assume that this entity has alternative possibilities for choice (List 2019, 2023). Here I go beyond Dennett's account of free will, which puts less emphasis on alternative possibilities.

Although intentional explanations are not always presented in a formally precise manner, an intentional explanation of an entity's behaviour typically proceeds as follows:

1. The explanation assumes that the entity has a choice between different possible options.
2. The explanation assumes that the entity somehow considers or evaluates these options from a goal-directed perspective.[12]
3. The explanation assumes that the entity chooses one of the options on that basis.

For example, when political scientists explain why people vote for a particular party, they assume that those people face a choice between different parties, consider them based on their preferences, and make an intelligible (albeit perhaps not always rational) choice on that basis.

[9] Retzlaff et al. (2024) further distinguish between "post-hoc" and "ante-hoc" explainability: "[t]hey are distinguished based on whether a model is intrinsically explainable (ante-hoc), or whether explainability is achieved by [an analysis of] the model after training (post-hoc)" (p. 2).

[10] Indeed, Retzlaff et al. (2024) distinguish explainability from what they call "interpretability", which they define as "understanding the inner workings and mechanisms of an AI model, seeking to answer questions like '*How does the model work?*'", as seen "from a developer or researcher's perspective" (p. 2). The latter might focus somewhat more on low-level mechanisms.

[11] Miller (2019) notes that "it is not a stretch to assert that people will expect explanations using the same conceptual framework [namely, belief-desire-intention psychology] used to explain human behaviours. This model is particularly promising because many knowledge-based models in deliberative AI either explicitly build on such folk psychological concepts, such as belief-desire-intention (BDI) models …, or can be mapped quite easily to them; e.g. in classical-like AI planning, goals represent desires, intermediate/landmark states represent intentions, and the environment model represents beliefs" (p. 17).

[12] "Consideration" or "evaluation" could be anything ranging from slow and deliberative to fast and instinctive.

Similarly, when economists explain consumer behaviour, they assume that consumers face choices between different consumption bundles, compare these options based on their preferences, and make a choice. The present explanatory scheme (consisting of steps 1 to 3) is at least implicit in intentional explanations across a wide range of academic fields. And clearly, this explanatory scheme could not work without the assumption that the relevant agents face choices: alternative possibilities are a presupposition of intentional explanations (on the nature of agentive possibilities, see also Maier 2015, 2022).

This point can be reinforced by noting that intentional explanations, whether in the social sciences or in artificial intelligence, effectively have a decision-theoretic format: they attribute choice options to the agents and a mechanism of choosing one option from amongst several possible ones. Thus, if we view an AI system as an intentional agent and explain its behaviour through a decision-theoretic lens – something central to many approaches to AI, including Russell and Norvig's (2021) – we must assume that the system has alternative possibilities to choose from, just as economists or political scientists assume that human agents choose between alternative possibilities (Maier 2023, List 2023).

Again, the desideratum of explainability favours systems that can be viewed as choosing between alternative possibilities. As noted by Miller (2019), good explanations in response to "why" questions are typically *contrastive*: "they are sought in response to particular counterfactual cases… That is, people do not ask why event P happened, but rather why event P happened instead of some event Q" (p. 3). For instance, we might ask: why did an agent do X rather than Y? Why did a particular self-driving car fail to stop at the junction rather than stop? Why did an autonomous trading system go ahead with a particular trade rather than refrain from doing so? Why did the diagnostic system arrive at a positive rather than negative diagnosis on some case? And so on. By attributing alternative possibilities to a system among which this system makes a choice, intentional explanations have the desired contrastive format.

*4.3. Causal control*

The third question we must ask is whether we have good reasons to think that AI systems have causal control over their actions. It is first helpful to clarify how one would understand "causal control" in the human case. The key question there is whether the person's high-level mental states, such as the intention to perform the action, make the right causal difference to that action such that the action is not exclusively explained by physical states of the brain and body, like a bodily reflex (Woodward 2008, List and Menzies 2009, Raatikainen 2010). For a mental state to be a *difference-making cause* of an action, in turn, two counterfactuals must be true:

1. If the person did not have that mental state, they would not perform the action.
2. If the person had that mental state in other similar circumstances, they would also perform the action.

For example, my intention to vote "yes" in a committee (a mental state) is the difference-making cause of the act of raising my arm at the right moment. If I didn't have that intention, I wouldn't raise my arm; and if I had the intention in other similar circumstances, I would also raise my arm. Thus a mental state is a difference-making cause of an action if the performance of the action systematically co-varies with the presence or absence of that mental state, holding other things fixed (List and Menzies 2009). The difference-making account of causal control can be spelt out further, but what matters for present purposes is that, in the case of genuine actions as opposed to mere bodily processes like digestion or reflexes, we cite mental states as the explanatorily significant difference-makers. Note that citing a mental state as a difference-making cause of an action is consistent with recognizing that this mental state is implemented by physical states of the brain and body.

A similar analysis can be given by introducing the notion of a "control variable" for some outcome. Control variables are "parameters which, when changed, lead to systematic changes in other variables of interest" (Roskies 2012, p. 329). When we explain human actions, as Campbell (2010, p. 26) notes,

"(a) psychological variables [such as mental states] function as control variables for the outcomes in which we are interested,
(b) what is going on at a psychological level of description supervenes on [is implemented by] what is going on at a physical level of description, but
(c) at the physical level, there are no control variables for the outcomes in which we are interested."

Physical-level states, such as highly specific neural states of the brain, are too fine-grained to serve as control variables for human actions. Recall again that we would explain the raising of my arm when a vote is taken by citing my voting intention rather than a particular microstate of my brain and body. Indeed, influencing human behaviour at the agential level, by providing people with information and motivations, is usually more effective than influencing it at a purely physical level, by trying to influence brain activity directly.

Now it should be clear what it would take for an AI system to have causal control over its actions. The key question is whether, to explain the system's actions, it is always better – more empirically adequate, more informative, more parsimonious – to cite low-level microstates of the system as causes, for instance descriptions at the level of the underlying algorithm, or whether it is sometimes better to cite high-level representational or goal states. Equivalently, we must ask whether the control variables for the system's actions are always to be found at a low algorithmic level or at least sometimes at a higher representational level. In the latter case, our explanations of what the AI system does would refer to the analogues of mental rather than physical causes. The system could then be said to have causal control over its actions.

Once more, the quest for explainable AI is relevant. An AI system whose actions systematically co-vary with its representational and goal states will be more explainable than one whose behaviour can be viewed only as the opaque result of low-level algorithmic processes. Explainability thus gives us a reason to design AI systems that meet the condition of causal control.

In sum, to the extent that some AI systems are best explained as intentional agents with alternative possibilities to choose from and causal control over their actions, those systems may be said to have free will under the present definition. Consistently with this, AI systems can still vary significantly in their agential complexity and sophistication. A system's intentional agency can be of a rich or narrow sort; the scope for choice between alternative possibilities can be richer or narrower; and the extent to which a system exercises control over its actions can also vary from system to system. Judgments of free will may thus come in degrees.[13]

## 5. Further questions

My argument for the possibility of free will in AI invites several further questions.

*5.1. If a computer implements an AI system that satisfies the above-mentioned conditions for free will, which entity is the bearer of free will: the computer, the software, or something else?*

It seems counterintuitive to suggest that a computer or a smartphone could acquire free will simply by running an appropriate AI app. On my account, however, free will is a high-level property of the AI system in its entirety, not a property of the underlying physical device. So, free will would be a system-level property of the choice-making agent that is being *implemented* by the computer with the relevant software. Free will is not a property of the computer *qua* physical device. In the same way, one would say that a human being has free will *qua* intentional agent, and not that the underlying brain has free will *qua* physical organ. The brain is part of the hardware that *implements* the high-level system with free will.

*5.2. Doesn't the fact that AI systems are based on deterministic algorithms rule out free will from the outset?*

The first thing to note is that, according to the widely held "compatibilist" view in philosophy, free will is compatible with determinism, whether it is determinism in the brain and body or determinism in the physical world as a whole. Dennett (1984, 2003) is one among many philosophers who hold such a view. In a recent survey of professional philosophers, almost 60% of respondents described themselves as compatibilists (Bourget and Chalmers 2023). If compatibilism is correct, the mere fact that an entity's "hardware" is deterministic would not exclude the possibility that the entity has free will. This point applies not only to human beings but arguably also to AI systems (Maier 2023).

[13] Floridi and Sanders (2004, p. 349) share the view that AI systems can qualify as agents in a high-level sense, noting that "[a]genthood … depends on a [level of analysis]", but they suggest that the relevant concept of agency might "not necessarily [exhibit] free will, mental states or responsibility". Yet they take "mindedness" and free will to require "some special internal states, enjoyed only by human and perhaps super-human beings" (p. 366). As noted, AI systems are unlikely to have such special internal states. However, once we understand agency and free will in the present, less metaphysically demanding way, there may be more common ground between their view and mine. Floridi and Sanders concede that some artificial agents "are already free in the sense of being non-deterministic systems" (p. 366), which I would understand as "non-deterministic at the relevant level of analysis", as discussed in section 5.2 below. The common ground continues when they say: "the agents in question satisfy the usual practical counterfactual: they could have acted differently had they chosen differently, and they could have chosen differently because they are interactive, informed, autonomous and adaptive" (p. 366).

However, since I have included alternative possibilities in the definition of free will, I cannot appeal to those versions of compatibilism that drop the alternative-possibilities requirement for free will. Recall that Dennett, for instance, drops this requirement in some of his writings (e.g., Dennett 1984). If we retain that requirement, then it is not clear how a system that is based on deterministic algorithms could make real choices between alternative possibilities. At any time, the system's state would fully determine what the system does next; the system would never face a genuine fork in the road, where it could do one thing or another.

However, as already noted, some compatibilists – including Dennett in other writings (e.g., Dennett 2003 and Taylor and Dennett 2002) – have offered strategies for (re)defining the notion of "having alternative courses of action" such that it becomes compatible with determinism. To explain my own preferred strategy, applied to the case of AI, let me begin by recalling that there are different levels at which we can describe an AI system: a micro-level at which we refer to (gazillions of) binary operations in logic gates, and a macro-level at which we refer to the cognitive and agentive processes realized. Micro-level descriptions are more fine-grained, macro-level ones more coarse-grained. Once we recognize that there are different such levels of description, we can also see that there are different notions of "possibility", which are relevant at different levels. At the micro-level the relevant notion is *possibility, conditional on the system's micro-state*, while at the macro-level it is *possibility, conditional on the system's macro-state*. The latter (macro-level) notion is less constrained than the former (micro-level) one, insofar as it conditionalizes on a coarse-grained macro-level state instead of a fine-grained micro-level state. This, in turn, implies that the more permissive, macro-level notion of possibility may admit alternative possibilities even when the more restrictive, micro-level notion doesn't. Consequently, our best *macro-level* explanations of a system may describe that system as indeterministic, even if its underlying *micro-level* processes are deterministic. In fact, this point holds generally for systems that can be described at different levels (Butterfield 2012, Yoshimi 2012, and List 2014): the determinism/indeterminism distinction is not preserved under changes in the level of description that we use to analyze a system.

On this account, what matters for alternative possibilities in any system – human, animal, AI – is whether the system is *best explained* by depicting it as an intentional agent capable of choosing between alternative possibilities. If the answer is "yes", then the fact that, at a lower level, there are deterministic processes is irrelevant. I propose that "alternative possibilities" in the context of free will should be understood as alternative possibilities at the (macro-)level of agency (following List 2014, 2023). And as already argued, we have good explanatory reasons for attributing such alternative possibilities to AI systems, insofar as they qualify as choice-making agents. (The claim that there is a distinct "agentive" notion of possibility has also been defended by Maier 2015, 2022.)

*5.3 Does free will in AI systems require that these systems are unpredictable?*

It is sometimes suggested that unless a system is sufficiently unpredictable, it could not have free will. (For discussion in the AI context, see, e.g., Krausová and Hazan 2013, Sanchis 2018, and Hadley 2025.) However, saying that free will requires alternative possibilities at the level of agency is not the same as saying that free will requires unpredictability. Free will, even with the capacity for choice between alternative possibilities, is consistent with predictability. An entity may be predictable, in particular, because it makes its choices for intelligible reasons.

Human beings are often quite predictable, but this is not because they lack free will but because they make their choices based on reasons that a predictor may understand. For example, I disprefer alcoholic drinks and therefore choose a non-alcoholic drink each time I go to a restaurant. My friends, who know my preferences, can predict those choices, but that doesn't mean that I do not make genuine choices in the first place or that I lack the capacity to choose otherwise. When I enter a restaurant I face a choice, where I could do one thing or another. Even if I predictably choose the non-alcoholic drink, the choice is up to me. What matters for free will is that the agent qualifies as an intentional agent with a capacity for making choices and exercising control over the resulting actions, not that the resulting choices are unpredictable.

*5.4. Wouldn't the present analysis have the counterintuitive implication that even simple optimizing algorithms have free will?*

Consider a chess-playing computer. For each possible configuration of the chessboard, the system considers all possible moves permitted by the rules of chess and chooses the move deemed best by some objective function, which encodes the algorithm's "goals". This description has a decision-theoretic format and might thus suggest that our chess computer is a (simple) choice-making agent and thereby in principle the sort of entity that has free will. This conclusion is counterintuitive, since we could equally explain the chess computer in non-agential terms. We can think of it as a deterministic system whose possible states are the possible configurations of the chessboard and whose state change rule is a deterministic function mapping each state to a unique next state, namely precisely the one that, under the earlier, choice-theoretic description, would have been described as "maximizing the value of the relevant objective function". This redescription makes no reference to agency, choice, or alternative possibilities and seems to explain the system's behaviour equally adequately.[14] A similar point could be made about other entities that admit both non-intentional and intentional explanations, such as a thermostat. A thermostat can be viewed as a mechanical device, but also as a rudimentary agent that "chooses" between activating and de-activating the heating, depending on whether it "believes" the actual temperature is too low or too high relative to some "desired" target temperature (see also Dennett 1987).

[14] However, one might argue that *if* we need to refer to the original objective function to define the state change rule, we have not genuinely eliminated the choice-theoretic format. On formal differences between intentional and non-intentional explanations, see also Orseau, McGregor McGill, and Legg (2018).

I suggest that we should *not* attribute free will to a system unless viewing that system as a choice-making agent is explanatorily clearly superior to not doing so. Since non-agential explanations are perfectly feasible for the chess computer and the thermostat, and by some standards simpler, it is unnecessary to view those systems as choice-making agents. It would be an over-interpretation to ascribe free will to them; we would be ascribing to them richer cognitive capacities than they plausibly have. By contrast, in the human case, the ascription of choice-making agency is often explanatorily indispensable, and so the attribution of free will seems warranted. If an AI system is so complex as to render non-intentional explanations practically infeasible, then the same could be said about such a system.

*5.5. Isn't the claim that AI systems can have free will challenged by the fact that many such systems do not take any initiatives by themselves and act only when prompted to do so?*

Present AI systems are often only very reactive agents: they take actions only in response to human prompts, and once they have completed any given task, they remain passive and resume their activity only when prompted again. A self-driving car, for instance, does nothing until instructed to drive to a particular destination. Similarly, many LLM systems produce outputs only in response to specific prompts. In light of this, Nyholm (2018, p. 1201) has argued that "we ought not to regard [AI systems] as acting on their own, independently of any human beings. Rather, the right way to understand the agency exercised by these machines is in terms of human–robot collaborations, where the humans involved initiate, supervise, and manage the agency of their robotic collaborators."

My claim, however, is only that whenever a system is in a phase of choice-making agency, it exhibits a form of free will, by satisfying the three above-mentioned conditions. Perhaps some systems go into such a phase only when activated by certain prompts and remain "on standby" for the rest of the time, so that their agency becomes periodically inactive, a bit like a hibernating animal whose agency is temporarily dormant.

Furthermore, we can imagine AI systems that take on temporally extended tasks involving long phases of choice-making agency. Imagine an autonomous military drone that is tasked with monotoring a coastline on a long-term basis and that is capable of evaluating and flexibly responding to various threat situations, including ones that aren't predefined. One can think of such a system as capable of taking initiatives. Similarly, robotic pets may be designed to be spontaneous and to take initiatives, while pursuing longer-term goals, such as companionship with a human being. The extent to which a system has the capacity to take initiatives lies on a continuum and depends on how complex, flexible, and long-term its objectives are.

*5.6. If an AI system has free will, does this imply that the system is also conscious?*

While some researchers (such as Tononi et al. 2022) see free will and consciousness as being connected, the mainstream approach in philosophy is to treat them as separate. Conceptually, as I see it, consciousness is neither necessary nor sufficient for free will. Free will, as defined here, requires intentional agency, alternative possibilities, and causal control over one's actions. Consciousness requires the presence of subjective experiences that one undergoes from a first-person perspective: there is something it is like to be a conscious subject, for that subject, as Thomas Nagel (1974) famously put it. Free will can be understood as a third-personal notion, while consciousness is inherently first-personal. On the present picture, a system could in principle qualify as having intentional agency, alternative possibilities, and causal control over its actions – and thereby as having free will – without subjectively experiencing anything. An example of such a system would be a "philosophical zombie", which is considered by many to be a logically coherent, albeit purely hypothetical scenario (Chalmers 1996; Dennett disagrees; see, e.g., Dennett 2005). Conversely, a system could in principle have subjective experiences without actively choosing and controlling anything. A hypothetical example could be some kind of passive "experience machine".

Empirically, of course, paradigmatic agents with free will, such as human beings or other complex animals, may also be conscious, and vice versa; Dennett would presumably agree with this claim. If that is right, however, it would establish a contingent, rather than conceptually necessary connection between free will and consciousness, especially in the biological world. A conceptually necessary connection would be established only if one could show that intentional agency itself requires consciousness. According to some philosophical views, consciousness is indeed necessary for agency and/or intentionality, but my preferred methodology is to keep our theories as modular as possible and to use "thin" definitions of key concepts, which do not rely on too many built-in assumptions. For this reason, I here understand free will as a third-personally describable phenomenon, and I do not assume that free agents must also be conscious (even though, in practice, many or most of them are).

Needless to say, free will should not be confused with the *experience* of free will. One could have that experience without really having free will (for instance, if free will were an illusion). And if there could be free will in a non-conscious entity, then there could also be free will without any experience of it. Recall once more that I have defined free will in a pragmatic and metaphysically not too demanding way, and so it is not out of the question that some AI systems could qualify as having free will without having conscious experiences.

*5.7. Does free will in AI imply that AI systems are capable of bearing moral responsibility?*

It is sometimes assumed that having free will implies the capacity to bear moral responsibility. Dennett, for instance, suggests that the "varieties of free will worth wanting" are those that underwrite moral responsibility (e.g., 1984). Conceptually, however, it is useful to distinguish between free will and moral responsibility. Free will, as I have defined it here, is primarily a descriptive and explanatory notion; moral responsibility is normative.

Free will, on the present account, is the capacity for intentional agency, for choice between alternative possibilities, and for causal control over the resulting actions. This three-part capacity is what makes the explanatory logic of choice-making agency applicable to an entity – nothing more and nothing less. It is not, by itself, defined by reference to anything moral. Moreover, free will, so understood, need not be restricted to human beings, but could be present in non-human animals too. In fact, it is plausible that the relevant three-part capacity is present not just in humans but also, for example, in other primates, such as the great apes.

By contrast, the capacity to bear moral responsibility requires a particularly rich form of agency, namely moral agency, which includes the capacity for moral cognition. Non-human animals largely lack that capacity, despite having the sort of agency required for bare free will. Free will is thus necessary but not sufficient for the capacity to bear moral responsibility. That said, the quest for *ethical* AI may be viewed as the quest for designing artificial *moral* agents, and AI systems with free will may eventually become candidates for the ascription of moral responsibility. In sum, free will, as I have understood it here, is a fundamental agential capacity without which the richer capacities required for moral agency could not get off the ground.

## 6. Concluding remarks

To determine whether an AI system has free will, we should not be asking: does the system exhibit some mysterious property, is it unpredictable, or are its algorithms indeterministic? Rather, we should be asking: is the system *best explained* as an intentional agent, with the capacity for choice between alternative possibilities, and causal control over its actions? If we have good grounds for applying the explanatory logic of choice-making agency to a system, it may be said to have free will in a practically relevant, non-mysterious sense. The system then has a kind of "free will worth wanting", as Dennett (1984) calls it, and satisfies a key necessary (albeit not by itself sufficient) condition for bearing moral responsibility. Anyone who considers it desirable for AI systems to function as moral agents should find this conclusion congenial.

My argument for the possibility of free will in AI resembles a similar argument in the case of another class of artificial agents: corporations and other organized collectives. The claim that such entities constitute intentional agents can also be defended either on interpretivist grounds, inspired by Dennett (see Tollefsen 2015), or on functionalist-realist grounds, by noting that suitably organized collectives satisfy the functional conditions for belief-desire agency, albeit based on a social as opposed to electronic hardware (List and Pettit 2011, List 2021). Indeed, social scientists commonly explain the behaviour of such entities through the lens of decision or game theory, by attributing to them the ability to choose between different options in a goal-directed and strategically rational manner. For example, the theory of the firm in economics represents firms and corporations as rational profit-maximizing agents.

The recognition that some collectives constitute choice-making agents in their own right raises the issue of free will too. At first, one might think that "corporations (and other highly organized collectives like colleges, governments, and the military) are effectively puppets, dancing on strings controlled by external forces", as Kendy Hess (2014, p. 241) notes. However, Hess argues that once we properly recognize how corporate agents function, we have good reasons to think that they "act from their own 'actional springs' … and from their own reasons-responsive mechanisms" (ibid.).[15] This, for Hess, supports the claim that "they act freely and are morally responsible for what they do" (ibid.).

Furthermore, one may reach this conclusion not only by reference to the criterion of reasons-responsiveness, as invoked by Hess, but also by reference to the checklist of conditions used here: intentional agency, the capacity for choice between alternative possibilities, and control over the resulting actions. Corporations and other suitably organized groups agents arguably satisfy these conditions too (List 2025). The argument for free will in corporate entities is therefore similar to the one in the case of AI.

Free will, we may conclude, is not restricted to human beings and other complex animals, but can in principle occur in non-biological agents too. Group agents and AI systems are two different examples of a similar phenomenon.

**Acknowledgements**

I am very grateful to Ibrahim Hansu, Levin Hornischer, Seth Lazar, Sven Nyholm, Silvia Milano, and two anonymous reviewers for comments and suggestions. I also gratefully acknowledge helpful feedback I received when I presented this work at a MINT Lab seminar, Australian National University, Canberra, April 2025, and at the conference "Social Ontology Faces the Future II: Artificial and Collective Agents", Koç University, Istanbul, June 2025. The paper was originally circulated as an online working paper in December 2024.

[15] Hess attributes the notion of "acting from one's own actional springs" to Haji (2006).